\documentclass[11pt]{article}

\usepackage{etoolbox}

\def\PaperType{preprint}        

\ifdefstring{\PaperType}{camera}{
\usepackage[final]{acl}
}{}
\ifdefstring{\PaperType}{preprint}{
\usepackage[preprint]{acl}
}{}
\ifdefstring{\PaperType}{review}{
\usepackage[review]{acl}
}{}

\usepackage{times}
\usepackage{latexsym}

\usepackage[T1]{fontenc}

\usepackage[utf8]{inputenc}

\usepackage{microtype}

\usepackage{inconsolata}

\usepackage{graphicx}
\usepackage{fontawesome}
\usepackage[cachedir=minted-cache]{minted}
\setminted{
    fontsize=\small,
    breaklines=true     
}

\usepackage{bm}
\usepackage[dvipsnames]{xcolor}
\usepackage{multirow}
\usepackage{amsmath}
\usepackage{amssymb}
\usepackage{mathtools}
\usepackage[inline]{enumitem}
\usepackage{graphicx}
\usepackage{booktabs}
\usepackage{tabularx}
\usepackage{pifont}
\usepackage{hyperref}
\usepackage{mdframed}
\usepackage{dashrule}
\usepackage{caption}
\usepackage{subcaption}
\usepackage{diagbox}
\usepackage{xspace}
\usepackage[most]{tcolorbox}
\usepackage{spverbatim}
\usepackage[figuresright]{rotating}
\usepackage{scalerel}
\usepackage{algorithm,algpseudocode}

\algnewcommand{\LineComment}[1]{\State \(\triangleright\) #1}

\DeclarePairedDelimiterX{\infdivx}[2]{(}{)}{%
  #1\;\delimsize\|\;#2%
}

\newcommand{\ours}{B\textsuperscript{3}D-RWKV\xspace}

\newcommand{\mycustomsize}{\fontsize{8.5pt}{9.5pt}\selectfont}
\newcommand{\mycustomsizetwo}{\fontsize{11.0pt}{12.0pt}\selectfont}

\ifdefstring{\PaperType}{review}{
\NewDocumentCommand{\qingyun}{ mO{} }{\textcolor{orange}{\textsuperscript{\textit{Qingyun}}\textsf{\textbf{\small[#1]}}}}
\NewDocumentCommand{\kelin}{ mO{} }{\textcolor{cyan}{\textsuperscript{\textit{KeLin}}\textsf{\textbf{\small[#1]}}}}
}{
\newcommand{\qingyun}[1]{}
\newcommand{\kelin}[1]{}
}

\title{Triplet-Block Diffusion RWKV}

\author{
    Ke Lin\textsuperscript{*} \\ {\mycustomsizetwo William \& Mary} \\ {\ttfamily \small \href{mailto:leonard.keilin@gmail.com}{leonard.keilin@gmail.com}}
    \And
    Yiyang Luo\textsuperscript{*} \\ {\mycustomsizetwo HKUST} \\ {\ttfamily \small \href{mailto:yluodq@connect.ust.hk}{yluodq@connect.ust.hk}}
    \And
    Zhaolong Su \\ {\mycustomsizetwo Cornell} \\ {\ttfamily \small \href{mailto:zs494@cornell.edu}{zs494@cornell.edu}}
    \AND
    Yunya Song \\ {\mycustomsizetwo HKUST} \\ {\ttfamily \small \href{mailto:yunyasong@ust.hk}{yunyasong@ust.hk}}
    \And
    Anyi Rao \\ {\mycustomsizetwo HKUST} \\ {\ttfamily \small \href{mailto:anyirao@ust.hk}{anyirao@ust.hk}}
}

\begin{document}
\maketitle

\ifdefstring{\PaperType}{review}{}{
    \let\thefootnote\relax\footnotetext{\textsuperscript{*}Equal Contribution.}
}

\begin{abstract}
Causal Transformer language models suffer from strictly sequential decoding and a quadratic per-step attention cost.
While linear-time causal models and discrete diffusion models each address these weaknesses, their integration remains inherently inconsistent: diffusion requires bidirectional attention, while causal models are unidirectional.
To unify these architectures, we propose \ours, a diffusion RWKV variant that integrates the model's $\mathcal{O}(L)$ inference efficiency with parallel, bidirectional discrete-diffusion through a \emph{triplet-block layout} method. 
\ours-7.2B reaches comparable accuracy on an 8-task suite versus existing models while significantly outperforming baselines in decoding throughput with an average of $\mathbf{1.6\times}$ speedup.
\ifdefstring{\PaperType}{review}{
Code will be released soon.
}{
Code is available at \url{https://github.com/leonardodalinky/B3D-RWKV}.
}
\end{abstract}

\section{Introduction}

Large language models (LLMs) have advanced rapidly under the dominance of the strictly causal Transformer architecture \citep{vaswani2017attention}, yet the left-to-right design of most modern decoders introduces two structural limitations: sequential decoding, which prevents parallelization, and quadratic attention costs, which make long-context inference expensive. 
These drawbacks have driven the development of alternative architectures designed to challenge the Transformer's dominance:
(1) Discrete-diffusion language models \citep{nie2025large,bie2025llada,ye2025dream,gong2024scaling} avoid sequential decoding, instead denoising token blocks in parallel using bidirectional attention~\cite{arriola2025block}.
(2) The RWKV family \citep{peng2023rwkv,peng2024eagle,peng2025rwkv} reformulates the classical Recurrent Neural Network (RNN) with attention-like channel mixing to obtain $O(L)$ inference at Transformer-level quality. 


This motivates us to combine these alternative architectures to improve generation efficiency over standard Transformers.
However, using a strictly causal backbone for diffusion language models presents an architectural mismatch: diffusion requires bidirectional attention, while causal models are unidirectional.



To achieve this combination, we introduce a \emph{triplet-block layout} method that converts a causal RNN-style language model into a block-diffusion language model without altering the backbone. 
Each logical generation block of size $B$ appears three times consecutively in a training sample: a masked copy $b_1$, an identical masked copy $b_2$ on which the denoising loss is computed, and a clean ground-truth copy $b_3$ that refreshes the recurrent state before the next block. Because the backbone model reads strictly left-to-right, the hidden state arriving at any masked position of $b_2$ has already absorbed every unmasked token of $b_1$, so $b_2$ gains pseudo-bidirectional access to its own unmasked context on a strictly causal model.

\begin{figure*}[t]
\centering
\includegraphics[width=\linewidth]{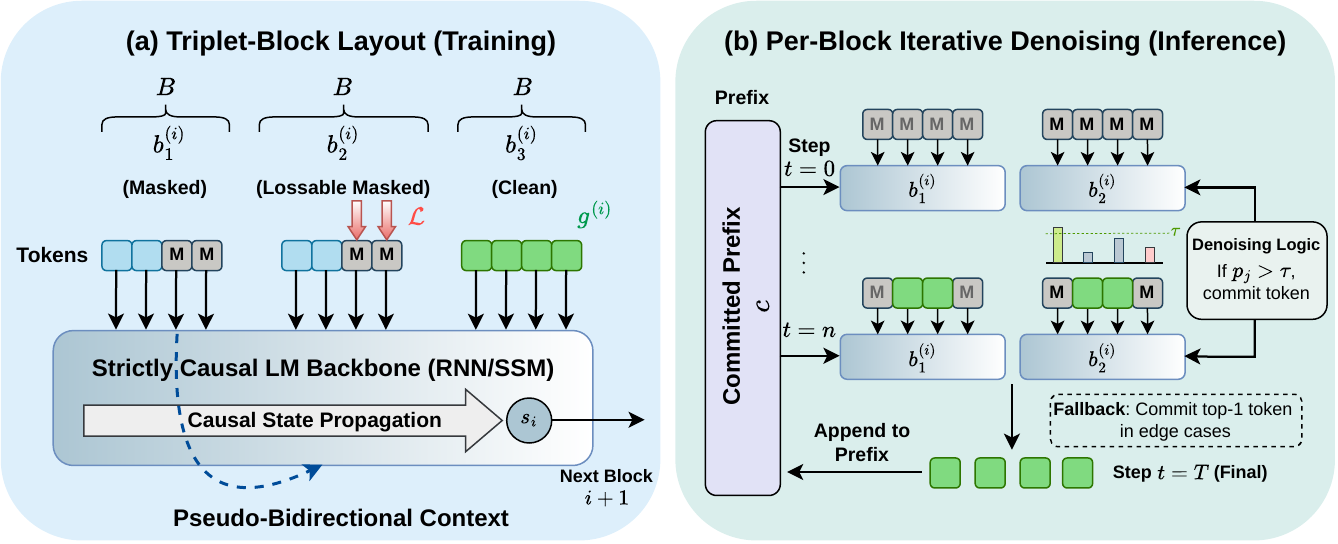}
\caption{
\textbf{(a)} Triplet-block layout for diffusion training on a strictly causal LM. Each logical block $i$ unfolds left to right as three contiguous physical blocks: a masked copy $b_1^{(i)}$, an identical lossable masked copy $b_2^{(i)}$, and a clean ground-truth copy $b_3^{(i)}$ that refreshes the recurrent hidden state for block $i+1$. 
\textbf{(b)} Per-block iterative denoising at inference. At each step, the sampler commits every position whose top-$1$ probability exceeds $\tau$. The loop terminates when every position is committed; the now-clean block is appended to $c$, and the next logical block begins.
}
\label{fig:fig_triplet_block_overview}
\end{figure*}

Our contributions are as follows:
\begin{itemize}
[leftmargin=8pt,itemsep=-4pt,topsep=0pt]
\item We release \ours -7.2B, the first diffusion-style linear-time RNN language model trained at the 7B scale with the mask-prediction objective. We train the model using our triplet-block diffusion framework, which integrates parallel token selection into the RWKV-7 backbone without modifying its original parameters. 
\item We provide a comprehensive comparison between our model and other strictly causal language models on an 8-task suite. We also demonstrate that our 7.2B model matches the reasoning capabilities of the RWKV-7 baseline while achieving $1.6\times$ the decoding throughput at comparable generation lengths.
\end{itemize}

\section{Related Work}
\label{sec:related}

\paragraph{Discrete-diffusion and masked language models.} The thread traces back to BERT-style masked-language pretraining \citep{devlin2019bert} and Mask-Predict's parallel decoder \citep{ghazvininejad2019maskpredict}, which MaskGIT \citep{chang2022maskgit} carried to image transformers with a confidence-thresholded commit schedule that almost every later masked generator reuses. The discrete-diffusion family proper was introduced by D3PM \citep{austin2021structured}, with SEDD \citep{lou2023discrete}, MDLM \citep{sahoo2024simple}, and MD4 \citep{shi2024simplified} reformulating and simplifying the absorbing-state objective. More recent scaled-up systems, including LLaDA \citep{nie2025large}, LLaDA~2.x \citep{bie2025llada}, Dream~7B \citep{ye2025dream}, DiffuLLaMA \citep{gong2024scaling},  Block Diffusion \citep{arriola2025block}, WeDLM \citep{liu2025wedlm}, and Nemotron-Labs-Diffusion \citep{fu2026nemotronlabsdiffusion}, combine these objectives with instruction tuning and parallel decoding. The concurrent DiffuMamba \citep{singh2025diffumamba} is the closest design point to ours and the only prior recipe that pairs a masked-diffusion objective with a linear-time backbone, but it does so by \emph{architecturally modifying} Mamba into a bidirectional block and so trains from scratch at the 1.3B scale on DCLM \citep{li2024datacomplm}. 

\paragraph{Linear-time recurrent and state-space backbones.} A parallel thread has produced strictly causal, linear-time alternatives to softmax attention: the RWKV family from RWKV-4 \citep{peng2023rwkv} through Eagle/Finch \citep{peng2024eagle} to RWKV-7 \citep{peng2025rwkv}; the selective state-space models (SSM) Mamba and Mamba-2 \citep{gu2023mamba,dao2024transformers}; RetNet \citep{sun2023retentive}; Gated Linear Attention \citep{yang2023gated}; and the Hyena Hierarchy \citep{poli2023hyena}. These backbones report perplexity parity with quadratic-attention Transformers at large wall-clock and memory savings; to our knowledge, none have been combined with a discrete-diffusion training objective at a large scale.


\begin{table*}[t]
\centering
\mycustomsize
\begin{tabularx}{\linewidth}{c cccccccc}
\toprule
\multirow{2}{*}{\textbf{Model}} & \multicolumn{5}{c}{\textbf{General Tasks}} & \multicolumn{3}{c}{\textbf{Math \& Science}} \\
\cmidrule(l){2-6} \cmidrule(lr){7-9}
 & \textbf{MMLU(5)} & \textbf{ARC-C(0)} & \textbf{ARC-E(0)} & \textbf{PIQA(0)} & \textbf{RACE(0)} & \textbf{GSM8K(8)} & \textbf{MATH(4)} & \textbf{GPQA(5)} \\
\midrule
\multicolumn{9}{c}{\textit{Causal LM}} \\
LLaMA3-8B & 66.6 & 53.6 & 81.1 & 79.8 & 41.9 & 78.9 & 41.1 & 35.5 \\
Qwen3-8B & 76.9 & 56.6 & 81.7 & 79.1 & - & 89.9 & 60.8 & 44.4 \\
RWKV-7-7.2B & 65.1 & 55.5 & 83.8 & 80.7 & 43.5 & 83.9 & 48.8 & 30.8 \\
\midrule
\multicolumn{9}{c}{\textit{Diffusion LM}} \\
LLaDA-8B & 65.9 & 47.5 & 71.8 & 74.8 & 38.7 & 70.9 & 30.7 & 30.4 \\
Dream-7B & 69.5 & 59.8 & 83.9 & 75.8 & 44.7 & 77.2 & 39.6 & 36.6 \\
\midrule
\multicolumn{9}{c}{\textit{Strictly Causal Diffusion LM}} \\
DiffuMamba & - & 28.3 & 49.1 & 62.6 & - & - & - & - \\
\ours-7.2B & \underline{64.8} & \textbf{61.6} & \underline{79.3} & \underline{73.5} & \textbf{49.7} & \underline{71.5} & 23.8 & \underline{25.6} \\
\bottomrule
\end{tabularx}
\caption{Benchmark results on the 8-task suite. 
The number of few-shot examples for each benchmark is indicated in brackets.
Results outperforming the RWKV baseline are \textbf{bolded}, while comparable results are \underline{underlined}.
}
\label{tab:main}
\end{table*}

\section{Method}
\label{sec:method}


To enable diffusion paradigm within strictly causal language models, we propose a triplet-block layout for efficient training and inference.
This method comprises a triplet-block layout (\S\ref{ssec:layout}) and a block-wise iterative denoising sampler (\S\ref{ssec:inference}). Implementation details of training and inference are provided in Appendix~\ref{app:mask} and~\ref{app:infer}.

\subsection{Triplet-block layout}
\label{ssec:layout}

Let the training context length be $L$, and the logical generation block size be $B$. We partition each training sample into $N = L / B$ contiguous \emph{logical blocks}. For each logical block index $i \in \{1, \dots, N\}$, denote the clean ground-truth tokens by $g^{(i)} \in \mathcal{V}^B$. 
Each logical block is then laid out as the concatenation of three physical blocks of length $B$ (Fig.~\ref{fig:fig_triplet_block_overview}(a):
\begin{equation}
\underbrace{b_1^{(i)}}_{\text{masked copy}} \;\Vert\; \underbrace{b_2^{(i)}}_{\text{masked copy (lossable)}} \;\Vert\; \underbrace{b_3^{(i)}}_{\text{clean copy}}.
\end{equation}
The two masked copies $b_1^{(i)}$ and $b_2^{(i)}$ are identical: they share the same mask pattern $m^{(i)} \in \{0,1\}^B$, replacing masked positions with \textsc{[mask]} and retaining $g^{(i)}$ elsewhere. The clean copy $b_3^{(i)}$ is also identical to $g^{(i)}$. Let $\ell^{(i)}_j\!\in\!\{0,1\}$ be the \emph{lossable} flag, $\pi(i,j)$ the physical position of the $j$-th token of $b_2^{(i)}$, and $p_{ij}(\cdot) \!\triangleq\! p_\theta(\cdot \mid x_{<\pi(i,j)})$ the next-token distribution there. Writing $\mathcal{S}\!=\!\{(i,j)\!:\!m^{(i)}_j\ell^{(i)}_j\!=\!1\}$ for the supervised positions and $N_{\mathrm{v}}\!=\!|\mathcal{S}|$, the training loss is the mean cross-entropy on $\mathcal{S}$:
\begin{equation}
\mathcal{L}_{\mathrm{CE}}(\theta) = -\frac{1}{N_{\mathrm{v}}}\!\!\sum_{(i,j)\in\mathcal{S}}\!\!\log p_{ij}\!\big(g^{(i)}_j\big).
\label{eq:loss}
\end{equation}
Following the Confidence-Aware Parallel training scheme of LLaDA-2.0~\citep{bie2025llada}, we further sharpen $p_{ij}$ on supervised positions that are \emph{already} correctly predicted, so that the inference-time threshold sampler (\S\ref{ssec:inference}) can commit more positions per denoising step. Let $\hat{g}^{(i)}_j\!=\!\arg\max_v p_{ij}(v)$ be the model's current top-1, $H(p)\!=\!-\!\sum_v p(v)\log p(v)$ the entropy, $\mathcal{C}\!=\!\{(i,j)\!\in\!\mathcal{S}\!:\!\hat{g}^{(i)}_j\!=\!g^{(i)}_j\}$ the gated subset, and $N_{\mathrm{c}}\!=\!|\mathcal{C}|$:
\begin{equation}
\mathcal{L}_{\mathrm{CAP}}(\theta) = \frac{1}{N_{\mathrm{c}}}\!\!\sum_{(i,j)\in\mathcal{C}}\!\!H\!\big(p_{ij}\big).
\label{eq:cap}
\end{equation}
The membership in $\mathcal{C}$ is computed without gradient, so the entropy flows only on the selected subset. The total objective is
\begin{equation}
\mathcal{L}(\theta) = \mathcal{L}_{\mathrm{CE}}(\theta) + \lambda_{\mathrm{CAP}}\,\mathcal{L}_{\mathrm{CAP}}(\theta).
\label{eq:loss_total}
\end{equation}


\paragraph{Pseudo-bidirectional access.}
Fix any masked position with block-local index $j\!\in\!\{0,\dots,B\!-\!1\}$ in $b_2^{(i)}$, whose physical position $\pi(i,j)$ sits after every token of $b_1^{(i)}$. Two complementary streams of context are visible there. \textbf{(i) Left context.}~Within $b_2^{(i)}$ itself, the unmasked tokens at indices $k\!<\!j$ lie to the left of $\pi(i,j)$ and supply the standard causal left context that a vanilla decoder would use. \textbf{(ii) Right context via $b_1^{(i)}$.}~Because $b_1^{(i)}$ has been processed in full \emph{before} $\pi(i,j)$ and carries the \emph{same} mask pattern $m^{(i)}$, its unmasked tokens at every block-local index $k$ are already absorbed into the hidden state at $\pi(i,j)$. 
The union of streams~(i) and~(ii) is exactly the set of unmasked tokens of the logical block, so position $j$ receives full bidirectional conditioning over block $i$ while the backbone still reads the sample strictly left-to-right (Fig.~\ref{fig:fig_triplet_block_overview}). 
Although the training context is $3\times$ the original length, the architecture of strictly causal models remains more computationally efficient than standard attention-based models.


\paragraph{Requirements and universal claims.} The construction in Eq.~\eqref{eq:loss} and the pseudo-bidirectional-access argument depend on only two properties of the backbone $p_\theta$:
\begin{itemize}[leftmargin=8pt,itemsep=-6pt,topsep=0pt]
    \item \textbf{(R1)} \emph{Strict causality}: the predictive distribution at any position depends solely on positions strictly to its left.
    \item \textbf{(R2)} \emph{Forward-propagating state}: an internal state that allows the predictive distribution at positions $b_2^{(i)}$ to access unmasked tokens from $b_1^{(i)}$. 
\end{itemize}
Every member of the linear-time backbone family currently in use, such as RWKV-v4 through v7, Mamba and Mamba-2, RetNet, Gated Linear Attention, and Hyena, satisfies \textbf{(R1)} and \textbf{(R2)} by construction. 
Standard causal Transformers also satisfy these, but their triple sequence-length cost is unattractive. Therefore, the triplet construction defines a universally no-architectural-change training recipe over the class of strictly causal backbones. 

\subsection{Inference: block-wise iterative denoising}
\label{ssec:inference}

At inference, the model generates one logical block at a time with only 2 replicates of physical blocks. Let $c$ denote the prefix of already-committed tokens. For each new block, the sampler initializes to an all-\textsc{mask} input of length $B$ and runs at most $T$ denoising iterations. At each iteration the sampler forwards $c$ concatenated with the current best guess of the block, reads the top-$1$ probability $p_j$ at each still-masked position, and commits any position with $p_j > \tau$; a low-confidence fallback commits the top-$k_{\min}$ positions whenever fewer than $k_{\min}$ clear the threshold, guaranteeing strictly positive progress per iteration. 
Appendix~\ref{app:infer} records the per-iteration loop and Figure~\ref{fig:fig_triplet_block_overview} summarizes it.


\section{Experiments}
\label{sec:experiments}

\subsection{Setup}
\label{ssec:setup}

The backbone is the public \texttt{RWKV-7-g1f-7.2B} \citep{peng2025rwkv} causal-LM checkpoint. 
Training data is the mixture of T\"ULU\,3 SFT dataset~\cite{lambert2024tulu} and curated trajectories of GLM-5.1 and Claude Opus 4.6. 
In the first training round, we set the triplet layout (§\ref{ssec:layout}) to $B=32$ and $N=64$, expanding $2{,}048$-token samples into $6{,}144$-token sequences for 1.8 epochs. In the second round, we increase the layout to $N=256$, expanding $8{,}192$-token samples into $24{,}576$-token sequences for 0.2 epochs. The $\lambda_\text{CAP}$ is set to 0.5.
The model is trained on $8\times$ H100 80GB SXM GPUs.
The full setup is in Appendix~\ref{app:implementation} and \ref{app:notes}. 

\subsection{Benchmarks}
\label{ssec:main}

We evaluate \ours-7.2B on an 8-task suite: MMLU~\cite{hendrycks2020measuring}, ARC-Challenge, ARC-Easy~\cite{clark2018think}, PIQA~\cite{bisk2019piqa}, RACE~\cite{lai2017race}, GSM8K~\cite{cobbe2021training} and MATH~\cite{hendrycks2021measuring} and GPQA~\cite{rein2023gpqa}.
For a fair comparison, we restrict baselines to backbones of comparable parameter scale released in roughly the same time window as RWKV-7.

Table~\ref{tab:main} presents downstream performance on general and math reasoning tasks. \ours performs comparably to other diffusion LMs of similar scale and matches the performance of the RWKV-7 baseline.
Notably, our method outperforms others on benchmarks like ARC-C and RACE, likely due to pseudo-bidirectional perception-enhancing reasoning capabilities. 
Conversely, parallel decoding may slightly reduce math reasoning accuracy, as these problems involve highly complex structures.
For example, MATH is graded by a LaTeX-level answer verifier that demands exact symbolic and numerical agreement, leaving no partial credit for minor local errors, which is exactly the failure mode that parallel decoding is most exposed to.

These results show that \ours achieves comparable or superior performance on simpler tasks, but experiences an acceptable drop on complex structural problems, likely due to parallel decoding issues of diffusion language models.

\subsection{Throughput}
\label{ssec:throughput}


\begin{figure}[t]
\centering
\includegraphics[width=\linewidth]{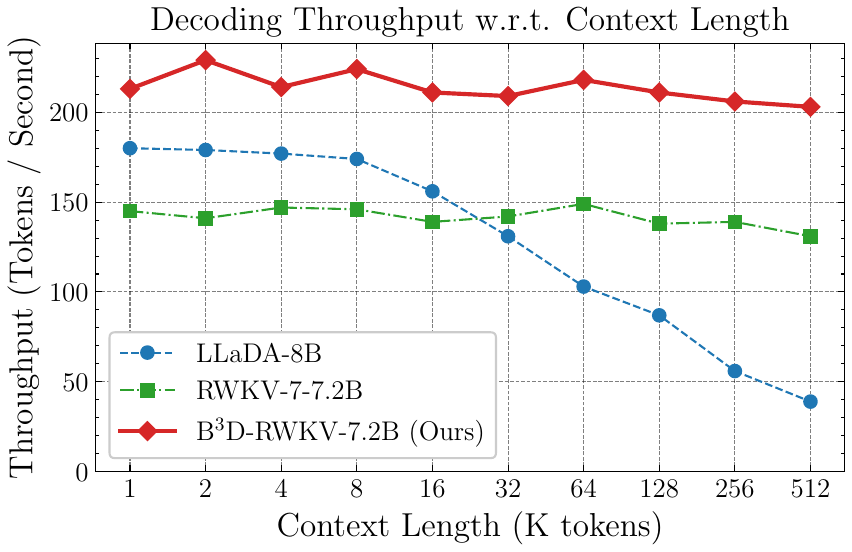}
\caption{
Inference throughput of LLaDA-8B, RWKV-7.2B, and \ours-7.2B on an H100 80GB GPU.
}
\label{fig:throughput}
\end{figure}

Figure~\ref{fig:throughput} compares the inference throughput of LLaDA-8B and our model against an RWKV-7 baseline across context lengths from 1K to 512K. LLaDA-8B utilizes \texttt{Fast-dllm}~\cite{wu2025fast} for efficient inference, with batch size fixed at 1, block size $B=32$, and $T=32$ diffusion steps. The commit threshold is set to 0.9 in our settings. Our model consistently achieves an average of $\mathbf{1.6\times}$ higher throughput than RWKV-7 while maintaining nearly identical performance. 
Adjusting sampling parameters achieves a 2.02× speedup with a slight drop in quality. More details of throughput is in Appendix~\ref{app:performance}.

\section{Conclusion}



We propose a triplet-block layout training method to adapt strictly causal language models into generative diffusion language models, and it requires no architectural changes.
\ours achieves a general $\mathbf{1.6\times}$ throughput compared to the original RWKV model, while maintaining comparable performance to existing models, offering an efficient way to transform pre-trained causal language models into diffusion language models.

\section*{Limitations}

\paragraph{Universality argued structurally; demonstrated on one backbone.} 
The universality claim is structural: the architectural requirements (R1, R2 in \S\ref{ssec:layout}) are stated precisely, and every member of the linear-time backbone family we cite satisfies them by construction. Due to computational constraints, we empirically validate the recipe using a single 7.2B-parameter RWKV-7 backbone. Empirical confirmation on smaller RWKV-v7 checkpoints, on other RWKV variants \citep{peng2023rwkv,peng2024eagle}, and on non-RWKV linear-time backbones, such as Mamba, is left to future work.

\paragraph{$3\times$ physical-sequence cost.}
The triplet layout applies a multiplicative $3{\times}$ factor to the physical sequence length per logical block in exchange for pseudo-bidirectional access. RWKV-7’s linear-in-length complexity \citep{peng2025rwkv} makes this approach feasible, whereas the quadratic cost of Transformers has previously restricted discrete-diffusion training to short context lengths.

\paragraph{Small-scale SFT data and no RL alignment.}
We continued training using only the T\"ULU 3 SFT mixture and a curated set of reasoning trajectories (Section 2), totaling 4.9B tokens. Additionally, we run neither large-scale further pretraining nor any subsequent reinforcement-learning alignment stage. Compared with the trillion-token corpus that produced the parent RWKV-7 ``Goose'' checkpoint, this is a narrow and stylistically biased distribution, so a degree of catastrophic forgetting on capabilities the base model originally acquired from broad pretrain data is essentially unavoidable, and likely accounts for part of the accuracy regression we observe on a subset of evaluation tasks. We expect that scaling the diffusion-style post-training corpus and adding an RL alignment stage on top of \ours would recover, and likely exceed, the parent checkpoint's accuracy on the affected tasks; both are left to future work.

\paragraph{More complicated scenarios.}
Due to computational constraints, we have not specifically optimized this model for scenarios like tool calling or coding. However, the pretrained RWKV’s inherent capabilities allow for success in some simple coding tasks, as demonstrated in Appendix \ref{app:samples}. We intend to improve this in future work.

\section*{Ethical Considerations}
\ours is initialized from a publicly released causal language model, and our fine-tuning only changes its architectural behavior, not its safety properties. We did not run any additional alignment or content filtering, so \ours inherits whatever biases and inaccuracies already exist in its base checkpoint and pre-training corpus. We recommend reviewing its outputs before using \ours in any user-facing system or in settings where factual accuracy matters.
Furthermore, the training dataset is publicly available and we do not impose any security check upon them.

\ifdefstring{\PaperType}{review}{}{
}

\bibliography{custom}

\clearpage

\appendix

\section{Model Implementation}
\label{app:implementation}
\subsection{Mask Sampling Rules}
\label{app:mask}

Following LLaDA \citep{nie2025large}, each logical block $i$ independently samples a mask ratio $r^{(i)} \sim \mathrm{Uniform}[0,1]$, then draws $\lfloor r^{(i)} B \rfloor$ mask positions uniformly without replacement from the lossable subset; with probability $0.10$ the ratio is overridden to $1.0$ (full-mask augmentation) so the training distribution matches the inference distribution, where each new block begins fully masked. Two role-aware rules are necessary and we treat them as non-negotiable: \emph{force-mask EOS} (the document-final \textsc{eos} is in every sample's mask set) and \emph{force-mask PAD inside the EOS-containing block} (trailing \textsc{pad} positions are extended to the mask). Both supervise the stopping decision and block a trailing-PAD shortcut that does not transfer to inference. 

\paragraph{Per-block uniform mask ratio.} Following LLaDA \citep{nie2025large}, each logical block $i$ independently samples its mask ratio $r^{(i)} \sim \mathrm{Uniform}[r_{\min}, r_{\max}]$ with $r_{\min} = 0$ and $r_{\max} = 1$ in all our runs. The mask pattern $m^{(i)} \in \{0,1\}^B$ is then drawn by sampling $\lfloor r^{(i)} B \rfloor$ positions uniformly without replacement from the lossable subset of block $i$.

\paragraph{Full-mask trick.} To shrink the distribution gap between training and inference, where each new block begins fully masked, we override $r^{(i)}$ to $1.0$ with probability $0.10$ at sample time, so roughly one in ten logical blocks has every lossable position masked. This is the LLaDA full-mask augmentation \citep{nie2025large} applied per-block rather than per-sample, and it is the only training-time hyperparameter we vary on the mask-sampling axis.

\paragraph{Force-mask EOS.} Every document carries a document-final \textsc{eos} token at $g^{(i^\star)}_{j^\star}$ for some $(i^\star, j^\star)$, where $i^\star$ is the EOS-containing logical block. We force-include this position in $m^{(i^\star)}$ in every sample of every epoch. Without this rule, a uniform per-block mask ratio of $r \sim \mathrm{Uniform}[0,1]$ supervises the EOS position in fewer than half of all samples and contributes only a small fraction of the total loss signal. Qualitatively, an unsupervised EOS produces an inference-time model that never stops generating, while force-masking the EOS yields well-calibrated stopping behavior.

\paragraph{Force-mask PAD inside the EOS-containing block.} A document shorter than $B$ tokens leaves a tail of \textsc{pad} symbols inside the EOS-containing logical block. Without intervention, the model learns the shortcut ``\textsc{mask} followed by visible \textsc{pad} implies \textsc{eos}'', a cue that does not transfer to inference because \textsc{pad} is suppressed in the decoder. We therefore extend the forced mask to all \textsc{pad} positions in the EOS-containing block. The model is then forced to predict the document-final \textsc{eos} from the upstream content rather than from a trailing-PAD shortcut.

\paragraph{Vocabulary slot reuse.} The RWKV-world tokenizer \citep{peng2025rwkv} uses $65{,}530$ real token slots out of its $65{,}536$-padded embedding table. We re-purpose two of the unused trailing slots, setting ID $65{,}535$ as \textsc{mask} and ID $65{,}534$ as \textsc{pad}, while preserving ID $0$ as the original \textsc{eos}. The embedding table and output projection are therefore not extended; no new parameters are introduced for the diffusion training objective.

\subsection{Inference Sampler}
\label{app:infer}

This appendix expands the per-iteration commit rule of \S\ref{ssec:inference}.

\paragraph{Per-iteration commit rule.} At each iteration $t$, the sampler forwards $c$ concatenated with the current best guess of the block through the backbone, reads the top-$1$ probability $p_j$ at each still-masked position $j$, and commits any position whose $p_j$ exceeds a fixed confidence threshold $\tau$. If \emph{fewer than} a floor $k_{\min}$ positions are committed in iteration $t$, the sampler falls back to committing the top-$k_{\min}$ most confident positions even when they sit below $\tau$. This guarantees strictly positive progress per iteration. The rule is the LLaDA~2.0 \citep{bie2025llada} confidence-threshold-plus-low-confidence-fallback rule, which itself descends from the MaskGIT \citep{chang2022maskgit} confidence-thresholded commit schedule for non-autoregressive masked generators. Iteration $t+1$ then conditions on the newly-committed tokens. The loop exits as soon as every position in the block is committed; the now-clean block is appended to $c$, and the next logical block begins.




\subsection{Architecture and Model Layout}
\label{app:arch}

Table~\ref{tab:arch} lists the architectural configuration of \ours-7.2B and the corresponding physical-sequence layout induced by the triplet block training scheme. All values match the public RWKV7-G1f-7.2B checkpoint we initialize from; we do not modify the backbone architecture.

\begin{table}[t]
\centering
\begin{tabularx}{\linewidth}{lr}
\toprule
\textbf{Metric} & \textbf{Value} \\
\midrule
Layers & 32 \\
Embedding dim & $4{,}096$ \\
Head size & 64 \\
Number of heads & 64 \\
RWKV LoRA rank & 128/128/96/480 \\
Total parameters & 7.2~B \\
Vocabulary slots (padded) & $65{,}536$ \\
Vocabulary slots (used) & $65{,}530$ \\
\textsc{mask} token ID & $65{,}535$ \\
\textsc{pad} token ID & $65{,}534$ \\
Context length (training) & $6{,}144$ \\
Block size $B$ & 32 \\
Logical blocks per sample & 64 \\
Raw content len. per sample & $2{,}048$ \\
\bottomrule
\end{tabularx}
\caption{Architecture and physical layout of \ours-7.2B. The triplet layout maps each $2{,}048$-token raw content sample to a $6{,}144$-token physical sequence consisting of $64$ logical blocks of size $B = 32$, each unfolded as three contiguous physical blocks $(b_1, b_2, b_3)$.}
\label{tab:arch}
\end{table}

\begin{figure*}[t]
\centering
\includegraphics[width=\linewidth]{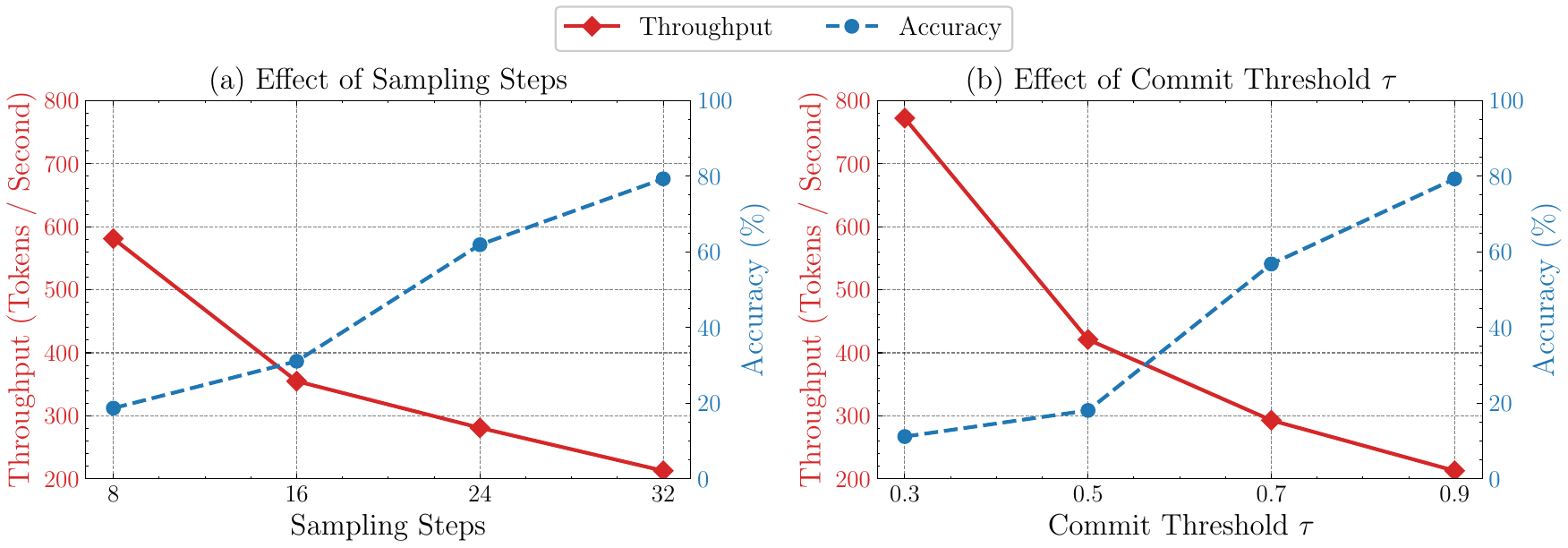}
\caption{
Effect of (a) sampling steps and (b) commit threshold $\tau$ on decoding throughput and accuracy on ARC-E benchmark.
}
\label{fig:ablation}
\end{figure*}

\section{Implementation Notes}
\label{app:notes}

This appendix records the precise distributed-training configuration used by our single \ours-7.2B run, for reproducibility. The notes here are not claimed as a contribution; they allow exact reproduction of the training run and flag two configuration pitfalls that are easy to overlook in the DeepSpeed \citep{rajbhandari2019zero} + PyTorch-Lightning stack.
Training takes a total of approximately 500 H100 hours using $8\times$ H100 80GB SXM GPUs to complete two epochs.

\paragraph{Training Dataset.} We trained the model on a 4.97 billion tokens mixture of the ``allenai/tulu-3-sft-mixture'', ``Jackrong/GLM-5.1-Reasoning-1M-Cleaned'', and ``angrygiraffe/claude-opus-4.6-4.7-reasoning-8.7k'' datasets on Huggingface Datasets.

\paragraph{Precision and gradient accumulation.} The run uses bf16 mixed-precision activations and a fp32 master copy of the optimizer state. Gradient accumulation precision must be set \emph{explicitly} to fp32 via the DeepSpeed configuration key $\mathtt{data\_types.grad\_accum\_dtype} = \mathtt{fp32}$. The Adam optimizer \citep{kingma2014adam} master weights, first-moment buffer, and second-moment buffer are all fp32 inside DeepSpeed's BF16\_Optimizer.

\paragraph{Inter-rank communication precision.} Inter-rank communication uses fp32 via $\mathtt{communication\_data\_type} = \mathtt{fp32}$ in the DeepSpeed configuration. Default communication-bucket sizes ($\mathtt{allgather\_bucket\_size}$, $\mathtt{reduce\_bucket\_size}$) are kept at their $200$ MB defaults; reducing them traded fragmentation for throughput in our exploratory tests.

\paragraph{Gradient clipping.} The PyTorch-Lightning Trainer's $\mathtt{gradient\_clip\_val}$ argument is silently ignored on the DeepSpeed strategy path. The same value must be propagated through the strategy's configuration as $\mathtt{strategy.config["gradient\_clipping"]}$ for it to take effect; we set this to $0.5$ matching the Trainer-side value, so both code paths agree.

\paragraph{Adam optimizer epsilon.} We set the Adam \citep{kingma2014adam} optimizer epsilon to $\epsilon = 10^{-8}$. The upstream RWKV training-code default of $10^{-18}$ is intended for the small-scale fp32 setting and is not appropriate for the GPU FusedAdam kernel that DeepSpeed's BF16 Optimizer dispatches to in our configuration; we therefore raise it to $10^{-8}$, the standard Adam default.

\paragraph{Parallelism.} We run DeepSpeed \citep{rajbhandari2019zero} ZeRO Stage 2 w/o offload on 8$\times$ NVIDIA H100 80GB GPUs. Effective batch size is $4$ batches per GPU $\times\,8$ GPUs $\times\,4$ gradient-accumulation steps $= 128$ samples per step.

\begin{table}[t]
\centering
\begin{tabularx}{\linewidth}{lc}
\toprule
\textbf{Component} & \textbf{Approx. peak (GB)} \\
\midrule
Model weights (bf16) & 14.4 \\
Gradients (bf16, ZeRO-2) & 1.8 \\
Adam state (fp32) & 10.8 \\
Activations & $\sim$30 \\
CUDA context overhead & $\sim$1.5 \\
\textbf{Total observed} & 58--60 \\
\bottomrule
\end{tabularx}
\caption{Single-GPU peak memory footprint (training-time) for the post-fix \ours-7.2B configuration on an H100 80GB. Activations and the wkv scratchpad dominate the budget.}
\label{tab:mem}
\end{table}

\paragraph{Memory footprint (Training).} Table~\ref{tab:mem} reports the observed single-GPU peak memory split for the post-fix configuration. The activation plus wkv-scratchpad budget dominates and is the binding constraint for raising the micro-batch beyond $4$ at the current context length.

\paragraph{Open-source License.} We follow the Apache-2.0 License from RWKV directly. 

\section{Performance}
\label{app:performance}

\paragraph{Throughput (Training).}
On 8$\times$ H100 80GB SXM GPUs, the post-fix training configuration sustains approximately $43{,}000$ tokens per second at micro-batch $4$ and context length $6{,}144$.

\paragraph{Throughput (Inference).}
Across $n=6{,}284$ requests sampled from a production-style workload, the mean decoding throughput is $222.1$ tok/s, with per-request rates ranging from $74.7$ to $785.4$ tok/s (Table~\ref{tab:throughput-stats}). The large standard deviation ($124.6$, about $56\%$ of the mean) comes mostly from differences in prompt length and in how many tokens the speculative decoder commits per draft step.

\begin{table}[t]
\centering
\begin{tabularx}{\linewidth}{X cccc}
\toprule
& \textbf{Avg} & \textbf{Std} & \textbf{Min} & \textbf{Max} \\
\midrule
Throughput & 222.1 & 124.6 & 74.7 & 785.4 \\
\bottomrule
\end{tabularx}
\caption{Per-request decoding throughput statistics over $n=6284$ requests (tokens/second).}
\label{tab:throughput-stats}
\end{table}

\begin{figure}[t]
\centering
\includegraphics[width=\linewidth]{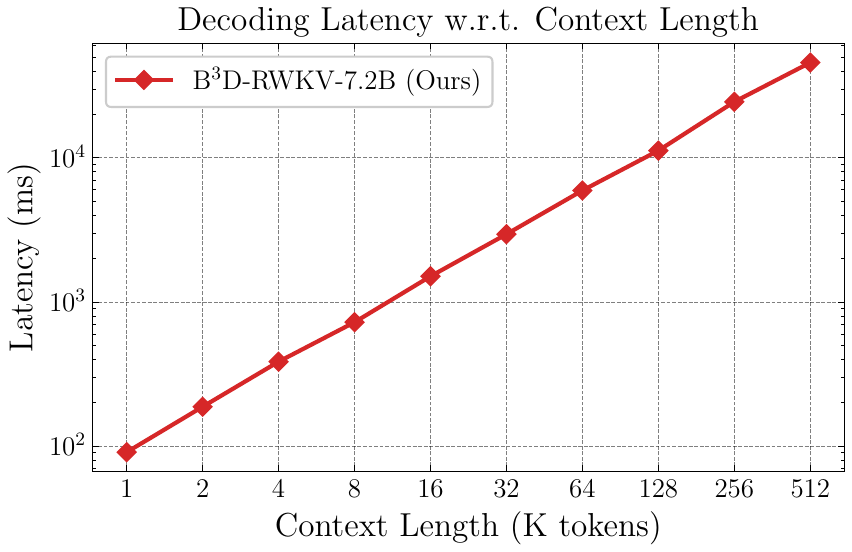}
\caption{
Latency in milliseconds by prefilled context length.
}
\label{fig:latency}
\end{figure}

\paragraph{Latency (Inference).}
End-to-end latency grows close to linearly with the prefilled context length, from $91$~ms at $1$K tokens to $45.8$~s at $512$K (Fig.~\ref{fig:latency}). An increase in context length produces a corresponding linear increase in latency, which is what we expect from the RWKV-7 backbone since it has no quadratic attention to dominate at long context.

\paragraph{Sampling steps.}
More sampling steps trade throughput for accuracy (Fig.~\ref{fig:ablation}~(a)). At $8$ steps, the model runs at $581$ tok/s but only reaches $18.7\%$ accuracy; at $32$ steps, accuracy climbs to $79.3\%$ while throughput drops to $213$ tok/s. The largest single jump in accuracy happens between $16$ and $24$ steps ($+30.8$ points for a $26\%$ throughput drop), which is roughly the point at which iterative denoising starts to produce coherent output.

\paragraph{Commit threshold.}
The commit threshold $\tau$ controls how confident the model must be before committing a draft token. At $\tau=0.3$ the decoder commits aggressively and reaches $772$ tok/s, but accuracy collapses to $11.2\%$ (Fig.~\ref{fig:ablation}~(b)). Raising $\tau$ to $0.9$ recovers $79.3\%$ accuracy at $213$ tok/s, which is the same operating point reached by $32$ sampling steps. Both knobs move along the same speed--accuracy frontier; we use $\tau=0.9$ as the default and treat lower values as a tuning knob when latency matters more than accuracy.

\section{Samples}
\label{app:samples}

We include several samples generated by \ours-7.2B in the Appendix to demonstrate the model's proficiency in solving both general and complex tasks. The default system prompt is the standard ``You are a helpful assistant''.

\section{Declaration of AI Usage}

We use Grammarly and Gemini to proofread our papers, and Claude Code for coding.

\definecolor{syscolor}{RGB}{128, 0, 128}       
\definecolor{usercolor}{RGB}{0, 85, 170}       
\definecolor{asstcolor}{RGB}{0, 128, 0}        
\definecolor{thoughtcolor}{RGB}{120, 120, 120} 
\definecolor{boxheader}{RGB}{50, 50, 50}       
\definecolor{boxbg}{RGB}{250, 250, 250}

\begin{table*}[t]
\begin{tcolorbox}[
    title={Sample: Knowledge}, 
    colback=boxbg,                          
    colframe=boxheader,                     
    coltitle=white,                         
    fonttitle=\bfseries,                    
    boxrule=0.7pt,                          
    arc=4mm,                                
    breakable=false 
]



\noindent\textcolor{usercolor}{\textbf{User:}} Could you tell me the life story of Albert Einstein?

\vspace{0.5em}

\noindent\textcolor{asstcolor}{\textbf{Assistant:}} 
{\color{thoughtcolor}\ttfamily \textless think\textgreater
The user is asking for a detailed biography of Albert Einstein's life. This is a straightforward request. I should cover his early life, his education, his scientific contributions, his challenges, and his later years. I need to present this in a clear, engaging way.

Let me think about what aspects of Einstein's life are most interesting. His early life in Munich, his time in Zurich, his work on the Einstein equation, his struggles with quantum mechanics, and his Nobel Prize are all key points. I should also mention his later work on unified field theory.

I should avoid getting bogged down in the technical details of his equations and focus on the human story. The tension of his relationship with quantum mechanics is a great story to include, as it shows his character and the limits of his work.

I'll structure this chronologically, starting from his birth in 1879 and going to his death in 1955. I'll include key events like his marriage, the Nobel Prize, and his later work.
\textless/think\textgreater} 

Albert Einstein (1879-1955) was one of the most influential physicists of the 20th century. Born in Munich, Germany, on March 14, 1879, he was the son of a Jewish family. His father, Hermann, was a successful manufacturer of electrical goods, and his mother, Pauline, was a seammaker.

Einstein showed an interest in mathematics from a young age.

\end{tcolorbox}
\end{table*}

\begin{table*}[t]
\begin{tcolorbox}[
    title={Sample: Biology}, 
    colback=boxbg,                          
    colframe=boxheader,                     
    coltitle=white,                         
    fonttitle=\bfseries,                    
    boxrule=0.7pt,                          
    arc=4mm,                                
    breakable=false 
]



\noindent\textcolor{usercolor}{\textbf{User:}} To prevent desiccation and injury, the embryos of terrestrial vertebrates are encased within a fluid secreted by the: A. Amnion, B. Chorion, C. Allantois, D. Yolk sac. Which is the right answer?

\vspace{0.5em}

\noindent\textcolor{asstcolor}{\textbf{Assistant:}} 
{\color{thoughtcolor}\ttfamily \textless think\textgreater
The user is asking about the embryonic sacs in terrestrial vertebrates. The correct answer is A. Amnion.

Let me think about this carefully. In terrestrial vertebrates, the embryo is enclosed in the amniotic sac, which contains the amniotic fluid. The amniotic fluid is produced by the amnion. The chorrion is the outermost layer of the embryonic sac, and the allantois is the waste absorption sac. The yolk sac is for nutrient absorption.

The chorrion is the outermost layer of the amniotic sac, and the fluid is the amniotic sac. The amnion is the membrane that produces the amniotic fluid. The chorrion is the outermost layer of the amniotic sac. The allantois is a sac that develops from the embryo.

So, the amnion is the structure that secretes the amniotic fluid.
\textless/think\textgreater} 

The correct answer is **A. Amnion**.

\end{tcolorbox}
\end{table*}

\begin{table*}[t]
\begin{tcolorbox}[
    title={Sample: Math}, 
    colback=boxbg,                          
    colframe=boxheader,                     
    coltitle=white,                         
    fonttitle=\bfseries,                    
    boxrule=0.7pt,                          
    arc=4mm,                                
    breakable=false 
]



\noindent\textcolor{usercolor}{\textbf{User:}} Joy can read 8 pages of a book in 20 minutes. How many hours will it take her to read 120 pages?

\vspace{0.5em}

\noindent\textcolor{asstcolor}{\textbf{Assistant:}} 
{\color{thoughtcolor}\ttfamily \textless think\textgreater
The user wants to find out how many hours it will take Joy to read 120 pages.

Joy reads 8 pages in 20 minutes.

First, find the number of pages she reads per minute.
Total pages = 8 pages in 20 minutes.
Pages per minute = 8 pages / 20 minutes = 0.4 pages per minute.

Next, find the number of minutes needed to read 120 pages.
Total pages = 120 pages.
Pages per minute = 8 pages.
Total minutes = 120 pages / 8 pages/minute = 15 minutes.
 
Then, convert minutes to hours.
15 minutes = 15 / 60 hours = 0.25 hours.

Let me double check.
Pages per minute = 8 / 20 = 0.4
Minutes per hour = 120 / 0.4 = 15 minutes.
Hours = 15 / 60 = 0.25 hours.

Alternatively: Pages per hour = 8 / 20 = 0.4 pages per minute.
Pages per hour = 8 / 20 = 0.4 pages per minute.
Pages per hour = 0.4 * 60 = 24 pages per hour.
Total hours = 120 / 24 = 5 hours.

Both methods yield 5 hours.
\textless/think\textgreater} 

It will take Joy **5** hours to read 120 pages. 








\end{tcolorbox}
\end{table*}

\begin{table*}[t]
\begin{tcolorbox}[
    title={Sample: Commonsense}, 
    colback=boxbg,                          
    colframe=boxheader,                     
    coltitle=white,                         
    fonttitle=\bfseries, 
    boxrule=0.7pt,                          
    arc=4mm,                                
    breakable=false 
]



\noindent\textcolor{usercolor}{\textbf{User:}} Where is the capital of US?

\vspace{0.5em}

\noindent\textcolor{asstcolor}{\textbf{Assistant:}} 
{\color{thoughtcolor}\ttfamily \textless think\textgreater
The user is asking for the capital of the United States. This is a straightforward geography question.
\textless/think\textgreater} 

The capital of the United States is Washington, D.C.

\end{tcolorbox}
\end{table*}

\begin{table*}[t]
\begin{tcolorbox}[
    title={Sample: Translation}, 
    colback=boxbg,                          
    colframe=boxheader,                     
    coltitle=white,                         
    fonttitle=\bfseries,                    
    boxrule=0.7pt,                          
    arc=4mm,                                
    breakable=false 
]



\noindent\textcolor{usercolor}{\textbf{User:}} Translate the following sentence to English: Hoy hace muy buen tiempo.

\vspace{0.5em}

\noindent\textcolor{asstcolor}{\textbf{Assistant:}} 
{\color{thoughtcolor}\ttfamily \textless think\textgreater
The user wants me to translate a Spanish sentence into English. The sentence is "Hoy hace muy bu tiempo."

"Hoy" = today, "hace" = makes, "muy buen" = very good, "tiempo" = "Hoy" = today, "weather" = weather

Translation: "It makes very good weather today."
\textless/think\textgreater} 

"It makes very good weather today.“

\end{tcolorbox}
\end{table*}

\begin{table*}[t]
\begin{tcolorbox}[
    title={Sample: Multi-turn}, 
    colback=boxbg,                          
    colframe=boxheader,                     
    coltitle=white,                         
    fonttitle=\bfseries,                    
    boxrule=0.7pt,                          
    arc=4mm,                                
    breakable=false 
]



\noindent\textcolor{usercolor}{\textbf{User:}} Calculate the result of 13 + 17.

\vspace{0.5em}

\noindent\textcolor{asstcolor}{\textbf{Assistant:}} The answer is **30**.

\vspace{0.5em}

\noindent\textcolor{usercolor}{\textbf{User:}} Now multiply the previous result by 3.

\vspace{0.5em}

\noindent\textcolor{asstcolor}{\textbf{Assistant:}} 
{\color{thoughtcolor}\ttfamily \textless think\textgreater
The user wants me to multiply the previous result by 3. The previous result was 30. 

30 * 3 = 90.
\textless/think\textgreater} 

The answer is **90**.

\end{tcolorbox}
\end{table*}

\begin{table*}[t]
\begin{tcolorbox}[
    title={Sample: Coding}, 
    colback=boxbg,                          
    colframe=boxheader,                     
    coltitle=white,                         
    fonttitle=\bfseries,                    
    boxrule=0.7pt,                          
    arc=4mm,                                
    breakable=false 
]



\noindent\textcolor{usercolor}{\textbf{User:}} Write a python function to calculate Fibonacci numbers.

\vspace{0.5em}

\noindent\textcolor{asstcolor}{\textbf{Assistant:}} 
{\color{thoughtcolor}\ttfamily \textless think\textgreater
\begin{verbatim}
We need to write a Python function to calculate Fibonacci numbers. 
The user says "Write a Python function to calculate Fibonacci numbers." 
That's straightforward. But we need to consider the context: The user 
likely wants a function that returns the nth Fibonacci number. Possibly 
they want an iterative or recursive approach. Since it's a simple request, 
we can provide a function that computes the nth Fibonacci number efficiently. 
We'll handle edge cases: n=0 returns 0, n=1 returns 1, maybe negative numbers? 
Typically Fibonacci is defined for non-negative integers. We'll assume n is 
a non-negative integer. We'll implement iterative for efficiency. We should 
output the code. The instruction: "Write a Python function to calculate 
Fibonacci numbers." So we just need to provide the function. Possibly include 
docstring and example usage.
We'll write:
def fibonacci(n):
    if n < 0:
        raise ValueError("n must be non-negative")
    a, b = 0, 1
    for _ in range(n):
        a, b = b, a + b
    return a
That's it.
\end{verbatim}
\textless/think\textgreater} 

\begin{verbatim}
```python
def fibonacci(n: int) -> int:
    """Return the nth Fibonacci number.
        The Fibonacci sequence is defined as:
            F(0) = 0
            F(1) = 1
            F(n) = F(n-1) + F(n-2) for n >= 1
        Args:
            n (int): A non-negative integer representing the position in the sequence.
        Returns:
            int: The nth Fibonacci number.
        Raises:
            ValueError: If n is negative."""
    if n < 0:
        raise ValueError("n must be non-negative")
    a, b = 0, 1
    for _ in range(n):
        a, b = b, a + b
        return a
```
\end{verbatim}

\end{tcolorbox}
\end{table*}

\end{document}